\documentclass[conference]{IEEEtran}
\IEEEoverridecommandlockouts
\usepackage{cite}
\usepackage{amsmath,amssymb,amsfonts}

\usepackage{graphicx}
\usepackage{textcomp}
\usepackage{xcolor}

\usepackage{hyperref}
\usepackage{algorithm}
\usepackage{algpseudocode}
\usepackage{subcaption}
\usepackage{tabularx,booktabs}
\usepackage{multirow}

\def\BibTeX{{\rm B\kern-.05em{\sc i\kern-.025em b}\kern-.08em
		T\kern-.1667em\lower.7ex\hbox{E}\kern-.125emX}}
\begin{document}

\title{MLP-Hash: Protecting Face Templates via \\Hashing of  Randomized Multi-Layer Perceptron
	\thanks{This research is based upon work supported by the H2020 TReSPAsS-ETN Marie Sk\l{}odowska-Curie early training network (grant agreement 860813). 
	}
}

\author{\IEEEauthorblockN{
		Hatef Otroshi Shahreza$^{1,2}$, Vedrana Krivoku{\'c}a Hahn$^{1}$, and S\'{e}bastien Marcel$^{1,3}$
	}
	\IEEEauthorblockA{$^{1}$Idiap Research Institute, Switzerland\\
		$^{2}$\'{E}cole Polytechnique F\'{e}d\'{e}rale de Lausanne (EPFL), Switzerland\\
		$^{3}$Universit\'{e} de Lausanne (UNIL), Switzerland}
}

\maketitle

\begin{abstract}
	Applications of face recognition systems for authentication purposes are growing rapidly. 
	Although state-of-the-art (SOTA) face recognition systems have high recognition accuracy, the features which are extracted for each user and are stored in the system's database contain privacy-sensitive information. Accordingly, compromising this data would jeopardize users' privacy.
	In this paper, we propose a new cancelable template protection method, dubbed MLP-hash, which generates protected templates by passing the extracted features through a user-specific randomly-weighted multi-layer perceptron (MLP) and binarizing the MLP output. 	
	We evaluated the  unlinkability, irreversibility, and recognition accuracy of our proposed biometric template protection method to fulfill the ISO/IEC 30136 standard requirements.
	Our experiments with SOTA face recognition systems on the MOBIO and LFW datasets show that our method has competitive performance with the BioHashing and IoM Hashing (IoM-GRP and IoM-URP) template protection algorithms.
	We provide an open-source implementation of all the experiments presented in this paper so that other researchers can verify our findings and build upon our work.
\end{abstract}


\section{Introduction}\label{sec:intro}
Face recognition has become a popular authentication tool and has been widely used in recent years.
The state-of-the-art (SOTA) face recognition systems mainly use convolutional neural networks (CNNs) to extract features, called ``embeddings'', from face images. 
In the enrollment stage, these features are extracted from each user's face and are stored as reference templates in the database of the face recognition system. 
Then, in the recognition stage, similar features are extracted from the user, and the resulting probe template is compared with the reference embedding stored in the system's database. 
These face features contain privacy-sensitive information about the user's identity \cite{mai2018reconstruction,template_inversion_icip2022}. 
Hence, data protection regulations, such as the EU General Data Protection Regulation (GDPR) \cite{GDPR}, consider biometric templates as sensitive data that must (legally) be protected.

To protect biometric templates, different methods have been proposed in the literature \cite{nandakumar2015biometric,sarkar2020review}. According to the ISO/IEC 30136 standard \cite{ISO30136}, each biometric template protection (BTP) scheme generally should have four main properties:
\begin{itemize}
	\item \textbf{Cancelability}: If a biometric template is compromised, we should be able to cancel the enrolled protected template and replace it with a new protected template.
	\item \textbf{Unlinkability}: 
	Considering the cancelability property, there should be no link between different protected templates from the same unprotected (original) biometric template.
	\item \textbf{Irreversibility}: It should be computationally difficult or impossible to recover the original biometric templates from the protected templates.
	\item \textbf{Recognition Accuracy}: The protected templates should allow for accurate recognition and should not result in recognition accuracy degradation.
\end{itemize}

BTP methods  can generally be categorized into \textit{cancelable biometrics} and \textit{biometric cryptosystems}. In cancelable
template protection methods (such as BioHashing \cite{jin2004biohashing}, Index-of-Maximum (IoM) Hashing \cite{jin2017ranking}, etc.) a transformation function is often used (which is dependent on a \textit{key}) to generate protected templates, and then for recognition the comparison is performed in the transformed domain \cite{nandakumar2015biometric, sandhya2017biometric}.
However, in biometric cryptosystems (such as fuzzy commitment \cite{juels1999fuzzy}, fuzzy vault \cite{juels2006fuzzy}, etc.), a key is either bound with a biometric template (called key binding schemes) or generated from a biometric template (called key generation schemes). Then, recognition is based upon correct retrieval or generation of the key \cite{uludag2004biometric}.

\begin{figure}[tb]
	\centerline{
		\includegraphics[width=1\linewidth,  trim={3.15cm 8.35cm 5.77cm 7.cm},clip]{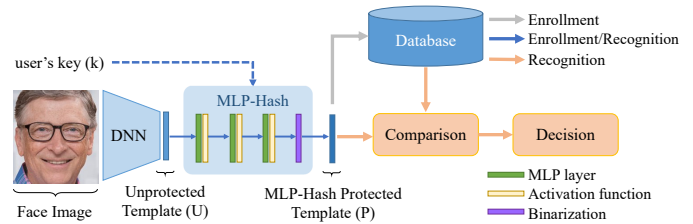}}\vspace{-1pt}
	\caption{Block diagram of MLP-Hash protected face recognition system}
	\label{fig:face_recognition_system}
	\vspace{-4pt}
\end{figure}

In this paper, we propose a new cancelable biometric template protection scheme, dubbed MLP-Hash, which includes a non-linear projection step through a user-specific randomly-weighted multi-layer perceptron (MLP), followed by a binarization step.  We employ the user's private key to initialize the MLP with  random orthonormal values. Then, we project the templates to a new space through the initialized MLP, which contains nonlinear activation functions. Finally, at the output layer, we binarize the final layer of the MLP  to generate the protected template. 

We evaluate  the unlinkability and irreversibility  properties  of our template protection method to fulfill the ISO/IEC 30136 standard  \cite{ISO30136} requirements. We also consider two scenarios when evaluating the method's recognition accuracy: the \textit{normal} scenario (which is the expected scenario in practice) and the \textit{stolen token}  scenario (which is the case when the user's MLP-Hash key is stolen). Then, we evaluate the protected templates of three SOTA face recognition methods (i.e., 
ArcFace \cite{deng2018arcface}, 
FaceNet \cite{FaceNet}, and 
InceptionResnetV2-CenterLoss \cite{de2018heterogeneous}) on the Labeled Faced in the Wild (LFW) \cite{LFW} and  MOBIO \cite{MOBIO} datasets. Our experiments show that MLP-Hash achieves promising performance in protecting SOTA face recognition systems.

The rest of this paper is organized as follows. First, we describe our biometric template protection method in section \ref{sec:porposed-method}. Then, in section \ref{sec:experiments}, we evaluate MLP-Hash in terms of unlinkability, irreversibility, and recognition accuracy. Finally, the paper is concluded in section \ref{sec:conclusion}.

\section{Proposed method}\label{sec:porposed-method} 
Figure \ref{fig:face_recognition_system} represents the block diagram of an MLP-Hash protected face recognition system. As depicted in this figure, MLP-Hash uses unprotected features, which are extracted from the user's face image, along with the user's key, to generate the protected template.  In section \ref{subsec:MLP-Hash}, we describe the MLP-Hash algorithm in detail. During the enrollment stage, the protected templates are stored in the system's database and are later compared with  the probe template during the recognition stage as described in section \ref{subsec:compare}. We should note that 
compared to BTP schemes which use neural networks and require training, e.g.~\cite{mai2020secureface,jang2019deep,lee2021softmaxout},  our proposed method does not require training and the weights are specified using the user's key (as described in section \ref{subsec:MLP-Hash}).

\subsection{MLP-Hash Algorithm}\label{subsec:MLP-Hash}
Let  $U$ indicate an unprotected biometric template (i.e., embedding) extracted by a face recognition model. The MLP-Hash protected template, $P$, can be generated by algorithm \ref{alg:1} using the user's key, $k$, and the unprotected template, $U$, in two steps. First, $U$ is fed into an MLP with $H$ hidden layers, activation function $F(.)$\footnote{In this paper, we use the Rectified Linear Unit (ReLU) activation function which is a non-linear and many-to-one function.}, and the pseudo-random orthonormal weights initialized with seed $k$. To generate pseudo-random orthonormal matrix  $\mathbf{M}_{\perp\ell}$ in  layer $\ell$ of the MLP, we first generate a pseudo-random matrix $\mathbf{M}_{\ell}$, and then apply the  Gram-Schmidt orthonormalization process on the rows of  $\mathbf{M}_{\ell}$. 
After feeding the $U$  into the MLP with the pseudo-random orthonormal weights, in the second step, we binarize the output of MLP to generate the protected template, $P$.

\begin{algorithm}[th]
	\small
	\caption{MLP-Hash algorithm}
	\label{alg:1}
	\begin{algorithmic}[1]
		\State \textbf{Inputs}: 
		\State \quad $U:$ unprotected biometric template (i.e., embedding)\\
		\quad	$H:$ number of MLP hidden layers\\
		\quad	$L_\text{MLP}:$ set of  lengths of MLP layers ($L_\text{MLP}^{(\ell)}$), including input layer ($\ell=0$), hidden layers ($1\leq \ell\leq H$), and output layer ($\ell=H+1$)\\
		\quad $F(.):$ activation function\\
		\quad $k:$ user's key
		\State \textbf{Output}:
		\State \quad $P=\{p_i | i=1,2,..., L_\text{MLP}^{(H+1)} \}$ binary MLP-Hash protected template
		\State \textbf{Procedure:}
		\State \quad \textbf{Step 1:} Passing through pseudo-random MLP
		\State\quad\quad Set initial value of $\Gamma$ with $U$ 
		\State\quad\quad \textbf{for} $\ell$ \textbf{in} $\{1, ...,  H+1\}$ \textbf{do}
		\State \quad\quad \quad Generate a pseudo-random matrix $\mathbf{M}_{\ell}$ based on the user's seed ($k$):  $\mathbf{M}_{\ell} \in \mathbb{R}^{L_\text{MLP}^{(\ell-1)} \times L_\text{MLP}^{(\ell)}} $.
		\State \quad\quad\quad  Apply the Gram-Schmidt process on the rows of   the generated pseudo-random matrix $\mathbf{M}_{\ell}$ to transform it into an orthonormal matrix $\mathbf{M}_{\perp\ell}$
		\State \quad\quad\quad  Update value of  $\Gamma$ with matrix product of $\Gamma$ and $ \mathbf{M}_{\perp\ell}$ 
		\State \quad\quad\quad  Update value of  $\Gamma$ by applying activation function $F(\Gamma)$
		\State\quad\quad \textbf{end for}
		\State \quad \textbf{Step 2:} Binarizing the output of MLP
		\State \quad\quad Compute ${L_\text{MLP}^{(H+1)}}$ bits MLP-Hash $\{p_i | i=1,2,...,{L_\text{MLP}^{(H+1)}}\}$ 
		
		from 
		$$p_i= \left\{
		\begin{array}{ll}
			0  & \mbox{if}\quad \Gamma_i \leq \tau \\
			1 & \mbox{if} \quad \Gamma_i > \tau
		\end{array}
		\right. ,\quad\quad i=1, ..., {L_\text{MLP}^{(H+1)}},$$ 
		\quad\quad where $\tau$ is the average of $\Gamma$ elements.
		\State \textbf{End Procedure}
	\end{algorithmic}
\end{algorithm}

\subsection{Comparing MLP-Hash Templates}\label{subsec:compare}
In the enrollment stage, the reference MLP-Hash templates, $P$, should be stored in the system database (ideally separately). 
In the recognition stage, we use \textit{Hamming} distance to calculate the score between each pair of \textit{probe} and \textit{reference} MLP-Hashed templates.
In the subsequent experiments, we consider the MLP-Hash protected face recognition systems operating in verification mode only.

\section{Experiments}\label{sec:experiments}
In this section, we describe our experiments and evaluate the properties of MLP-Hash as a biometric template protection scheme in accordance with the ISO/IEC 30136 standard. First, in section  \ref{subsec:exp_setup}, we describe our experimental setup and the baselines used. 
Next, we evaluate the  unlinkability, irreversibility and recognition accuracy of MLP-Hash in sections \ref{subsec:unlinkability}, \ref{subsec:irreversibility}, and \ref{subsec:recognition_performance}, respectively.
{We should note that cancelability  is inherently satisfied in the MLP-Hash algorithm, since like other \textit{cancelable} BTP methods, we can easily revoke the compromised template in the database, assign a new key for the user, and register the user with a new protected template.}
Finally, we discuss our experiments in section \ref{subsec:discussion}.

\subsection{Experimental Setup and Baselines}\label{subsec:exp_setup}
As stated in section \ref{sec:intro}, in our experiments we used the MOBIO  \cite{MOBIO} and Labeled Faced in the Wild (LFW) \cite{LFW} databases to evaluate the recognition accuracy of MLP-Hash on SOTA face recognition models.
The MOBIO dataset is a bimodal dataset including audio and face data acquired using  mobile devices from 152 people. We used the \textit{development} 
subset of the \textit{mobio-all} protocol\footnote{ The implementation of the  \textit{mobio-all} protocol for the MOBIO dataset is available at  \url{https://gitlab.idiap.ch/bob/bob.db.mobio}} in our experiments.
The LFW database includes 13,233 images of 5,749 people, where 
1,680 people have two or more images. We used the \textit{View 2} protocol\footnote{ The implementation of  the \textit{View 2} protocol for the LFW dataset is available at  \url{https://gitlab.idiap.ch/bob/bob.db.lfw}} to evaluate the models.
We also used three SOTA face recognition models\footnote{ The implementation of each face recognition model is available at  \url{https://gitlab.idiap.ch/bob/bob.bio.face}}, 
including 
ArcFace \cite{deng2018arcface}, 
FaceNet \cite{FaceNet}, and 
InceptionResnetV2-CenterLoss \cite{de2018heterogeneous}. We compare the performance of our template protection method on the same face recognition systems with the BioHashing \cite{jin2004biohashing} method and two methods based on Index-of-Maximum (IoM) Hashing \cite{jin2017ranking}   (i.e., Gaussian random projection-based hashing, shortly GRP, and uniformly random permutation-based hashing,  shortly URP).  
In each case, we generate protected templates whose length is equal to the length of the embedding (i.e., number of elements in the embedding) for each face recognition model. 
We set all the hidden layers of MLP-Hash to twice the length of the embeddings for each face recognition model. 
The number of hidden layers (H) was 3 in our experiments.

For our experiments, we used the Bob\footnote{Available at  \url{https://www.idiap.ch/software/bob/}} toolbox \cite{bob2012,bob2017}. To implement  the  BioHashing algorithm, we used the open-source  implementation of the BioHashing in Bob \cite{hatef_TBIOM2021,shahreza2021recognition}. 
The source code from our experiments is publicly available to help  reproduce our results\footnote{Source code:  \url{https://gitlab.idiap.ch/bob/bob.paper.eusipco2023_mlphash}}.

\begin{figure}[tb]
	\centering
	\begin{subfigure}[b]{0.24\textwidth}
		\centering
		\includegraphics[page=1,width=.95\linewidth, trim={0.7cm 0.5cm 1.7cm 0cm},clip]{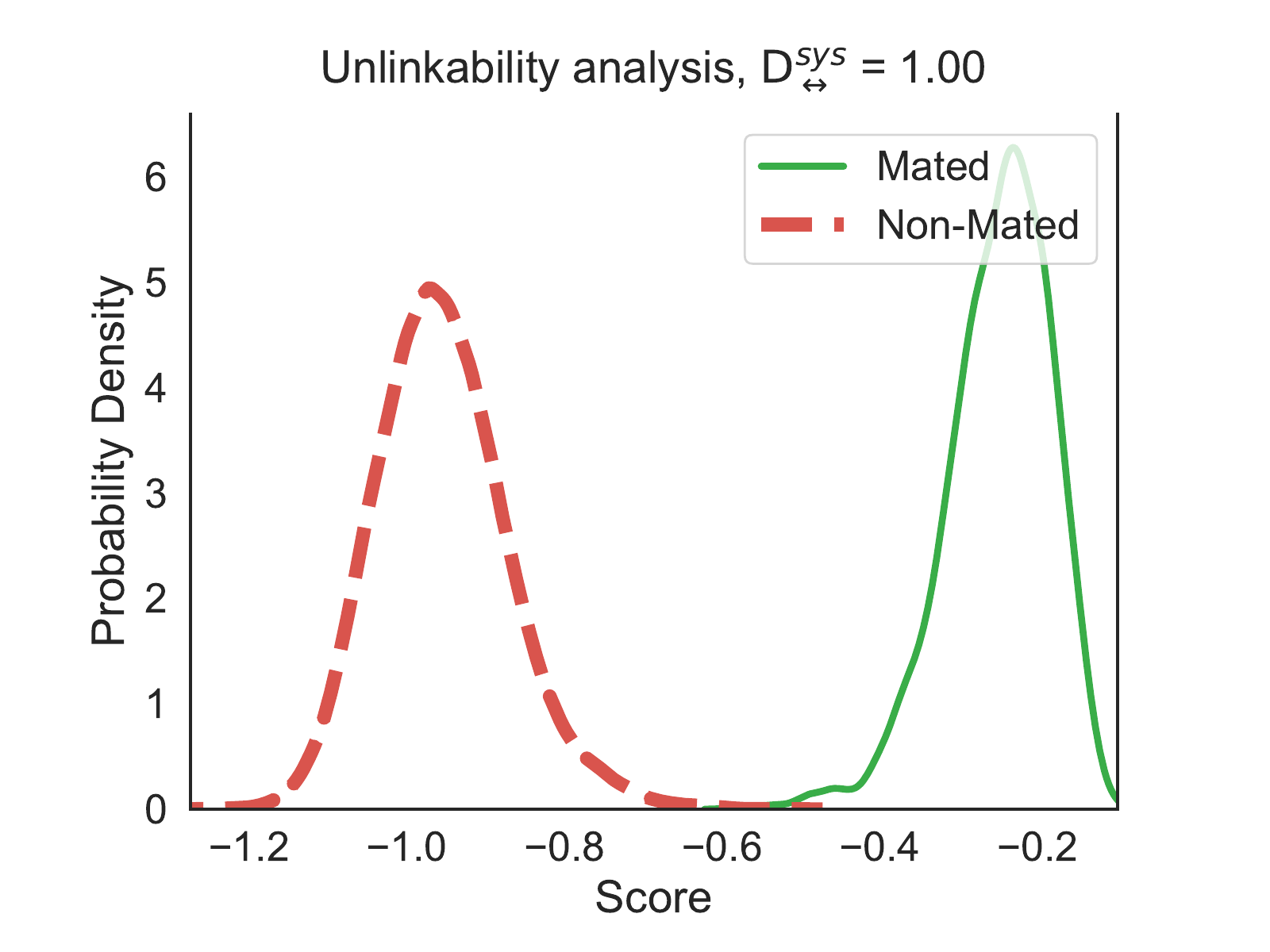}
		\caption{}
	\end{subfigure}\hfil
	\begin{subfigure}[b]{0.24\textwidth}
		\centering
		\includegraphics[page=1,width=0.95\linewidth, trim={0.7cm 0.5cm 1.7cm 0cm},clip]{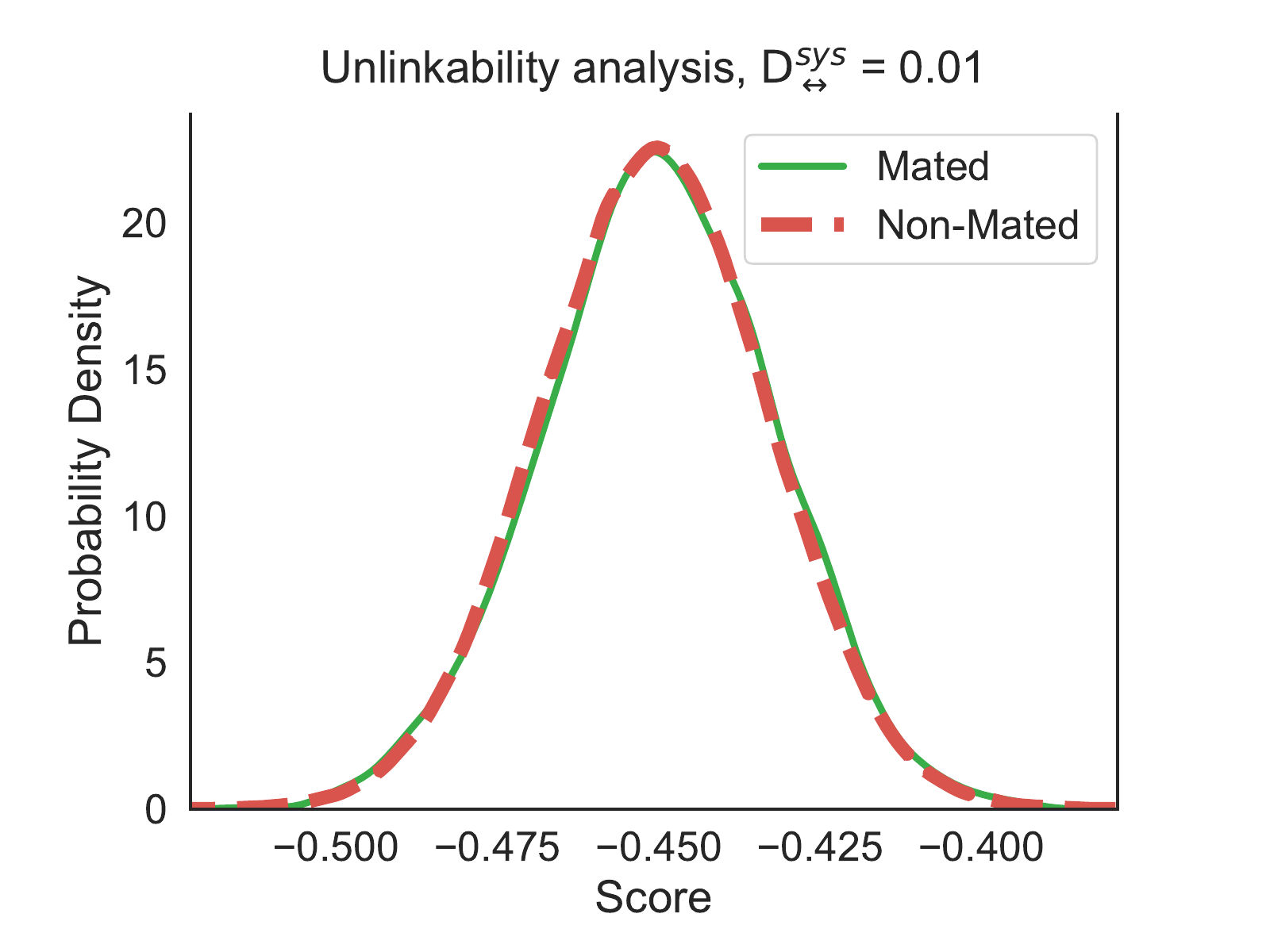}
		\caption{}
	\end{subfigure}\hfil \\\vspace{-5pt}
	\caption{Unlinkability evaluation of unprotected  and MLP-Hash protected ArcFace templates on  the MOBIO dataset: a)Unprotected templates, b)MLP-Hash protected templates.
	}\vspace{-9pt}
	\label{fig:unlinkability}
\end{figure}

\subsection{Unlinkability Evaluation}\label{subsec:unlinkability}
To evaluate the unlinkability criterion, we used the framework proposed in  \cite{gomez2017general}. 
{This framework uses the score distributions of the \textit{mated} templates (i.e., different templates from the same user) and \textit{non-mated} templates  (i.e., templates from different users) to measure unlinkability with respect to the overlap of these two distributions.
	More particularly, with this evaluation, we expect that in the case of linkable templates, the \textit{mated} and \textit{non-mated} templates score distributions will be separated. However, in the case of unlinkable templates, these distributions should completely overlap. }
Figure \ref{fig:unlinkability} compares the unlinkability  of original (unprotected) and MLP-Hash protected ArcFace templates on the MOBIO dataset using this evaluation framework\footnote{The corresponding plots for other models are also available in the software package of the paper.}.
{To calculate the distribution of \textit{mated} scores in this figure, we generated different templates for the same user using different keys, then calculated the scores between these templates. However, for the distribution of \textit{non-mated} scores, we generated protected templates for different users (with different keys) and computed the scores between them.}
As shown in this figure, while the distributions of \textit{mated} scores and \textit{non-mated} scores  are fully separated for unprotected templates, they  almost completely overlap for the MLP-Hash protected templates. Furthermore, the value of 
the system's global unlinkability  measure ($\mathrm{D}_{\leftrightarrow}^{\mathit{sys}}$)  
is reduced from $1.0$ (for the unprotected system) to $0.01$ (for the MLP-Hash protected system) by deploying our template protection method, showing that the resulting protected templates are almost fully unlinkable.   
Table \ref{tab:unlinkability} compares the unlinkability of  MLP-Hash, BioHash, IoM-GRP, and IoM-URP  protected templates  of  the  ArcFace embeddings on the MOBIO database. As this table shows, all these template protection schemes have comparable unlinkability and they are almost fully unlinkable.

\begin{table}[tbp]
	\centering
	\setlength{\tabcolsep}{3pt}
	\caption{Unlinkability evaluation of  MLP-Hash, BioHash, IoM-GRP, and IoM-URP  protected templates of the ArcFace embeddings in terms of the system's global unlinkability  measure ($\mathrm{D}_{\leftrightarrow}^{\mathit{sys}}$).}
	\label{tab:unlinkability}\vspace{-1pt}
	\scalebox{.95}{
		\begin{tabular}[t]{@{} cccc @{}}
			\toprule[1pt]
			{\textbf{MLP-Hash}} & {\textbf{BioHash}}& {\textbf{IoM-GRP}} & {\textbf{IoM-URP}} \\
			\midrule[1pt]
			0.010 & 0.009 & 0.011 & 0.007  \\
			\bottomrule[1pt]
		\end{tabular}
	}
\end{table}

\begin{table}[tbp]
	\centering
	\setlength{\tabcolsep}{3pt}
	\caption{Irreversibility evaluation of  MLP-Hash, BioHash, IoM-GRP, and IoM-URP  protected templates of the ArcFace embeddings in terms of Success Attack Rate (\%)  on the MOBIO dataset at FMR of $10^{-2}$ and $10^{-3}$.}
	\label{tab:irreversibility}\vspace{-1pt}
	\scalebox{.95}{
		\begin{tabular}[t]{@{} lcccc @{}}
			\toprule[1pt]
			{\textbf{Configuration}}   & {\textbf{MLP-Hash}} & {\textbf{BioHash}}& {\textbf{IoM-GRP}} & {\textbf{IoM-URP}} \\
			\midrule[1pt]
			\textbf{FMR = }$\mathbf{10^{-2}}$ &  39.05 & 43.81 & 35.71 & 14.29  \\ \midrule 
			\textbf{FMR = }$\mathbf{10^{-3}}$ &  9.05 & 10.48 & 7.14 & 1.43  \\  
			\bottomrule[1pt]
		\end{tabular}
	}
	\vspace{-2pt}
\end{table}

\begin{table*}[tbp]
	\centering
	\setlength{\tabcolsep}{3pt}
	\caption{Comparison of MLP-Hash-protected, BioHash-protected, IoM-GRP-protected, IoM-URP-protected, and unprotected (Baseline) SOTA Face Recognition models, in terms of TMR (\%)  in the \textit{normal} and the \textit{stolen} scenarios on the MOBIO and LFW datasets. The threshold in each system is selected individually at an FMR of $10^{-3}$. The results are reported as (mean$\pm$std) for 10 different experimental trials.}
	\label{tab:performance}\vspace{-1pt}
	\scalebox{.925}{
		\begin{tabular}[t]{@{} llcccccccccc @{}}
			\toprule[1pt]
			\multirow{2}{*}{\textbf{Dataset}} & \multirow{2}{*}{\textbf{Model}}  &
			\multirow{2}{*}{\textbf{Baseline\hspace{2pt}}}  & \multicolumn{4}{c}{\textbf{\textit{normal} scenario}} &&   \multicolumn{4}{c}{\textbf{\textit{stolen} scenario}}\\
			\cline{4-7}
			\cline{9-12}
			\rule{0pt}{2.5ex}    
			& &{} & {MLP-Hash} & {BioHash} & {IoM-GRP} & {IoM-URP} && {MLP-Hash} & {BioHash} & {IoM-GRP} & {IoM-URP}  \\
			\midrule[1pt]
			\multirow{3}{*}{\textbf{MOBIO}} 
			&ArcFace &  $100.00$ & $100.00\pm0.00$ & $100.00\pm0.00$  & $100.00\pm0.00$  & $99.59\pm0.08$ && $100.00\pm0.00$ &$99.95\pm0.04$  & $99.98\pm0.03$  & $98.88\pm0.13$  \\ \cline{2-12} \rule{0pt}{2.5ex}
			&FaceNet &   $97.87$ & $99.05\pm0.48$ & $99.93\pm0.04$  & $99.99\pm0.01$  & $95.56\pm0.50$ && $76.40\pm6.19$ &$89.38\pm2.12$ & $94.41\pm0.83$  & $87.51\pm1.03$    \\ \cline{2-12} \rule{0pt}{2.5ex}
			&IncResNetV2 &   $96.69$ & $99.98\pm0.04$ & $99.99\pm0.01$  & $100.00\pm0.00$  & $99.96\pm0.03$ && $65.12\pm5.07$ &$76.46\pm7.49$  & $92.56\pm2.73$  & $91.38\pm1.35$ \\ 
			\midrule
			\multirow{3}{*}{\textbf{LFW}} 
			&ArcFace &  $98.73$ & $98.86\pm0.13$ & $98.84\pm0.05$  & $99.19\pm0.06$  & $88.78\pm1.37$ && $95.54\pm0.54$ &$98.56\pm0.06$   & $98.62\pm0.06$  & $84.79\pm1.98$ \\ \cline{2-12}\rule{0pt}{2.5ex}
			&FaceNet &   $93.17$ & $90.90\pm0.90$ & $96.81\pm1.12$  & $99.38\pm0.09$  & $69.29\pm2.99$ && $59.42\pm5.02$ &$83.19\pm5.32$  & $85.78\pm4.92$  & $50.77\pm6.96$  \\ \cline{2-12}\rule{0pt}{2.5ex}
			&IncResNetV2 &   $93.33$ & $99.42\pm0.44$ & $99.95\pm0.05$  & $100.00\pm0.00$  & $98.06\pm0.35$ && $ 47.40\pm14.22$ &$64.13\pm19.89$ & $83.38\pm6.27$  & $81.38\pm2.75$  \\ 
			\bottomrule[1pt]
		\end{tabular}
	}\vspace{-2pt}
	
\end{table*}

\subsection{Irreversibility Evaluation}\label{subsec:irreversibility}
To evaluate the irreversibility of the proposed template protection scheme, we consider the worst-case and most difficult threat model in ISO/IEC 30136 standard (referred to as \textit{full disclosure threat model}), where the attacker knows everything about the system, including algorithms, secret keys, etc. We assume that the attacker would invert the protected template,  then use the inverted template to enter a similar unprotected system. Accordingly, we evaluate the irreversibilty in term of Success Attack Rate (SAR), which indicates the attacker's success rate in entering the unprotected system using the inverted templates. Hence, a higher SAR shows that the templates are more invertible, while a lower (or zero)  SAR indicates that the protected templates are harder to invert.

To evaluate such an attack, similar to \cite{hahn2022towards}, we used a numerical solver (implemented in the SciPy package\footnote{\url{https://scipy.org/}}) to find an estimate of the original template, which is mapped to the same output through the template protection module. 
The solver starts from an initial guess, and through an iterative process, tries to find an answer which gives the same output (as the given protected template) when passed as the input to the MLP-Hash with the same key. We  also assumed that the attacker knows the distribution of unprotected templates, and uses this distribution to extract 10 samples as initial guesses in separate attempts. In each attempt, in the case of convergence of the solver,  the inverted template is used to enter an unprotected system with a match threshold at a False Match Rate (FMR) of $10^{-3}$ (using the same feature extraction module). 

Table \ref{tab:irreversibility} compares the irreversibility of  MLP-Hash, BioHash, IoM-GRP, and IoM-URP  protected templates  of  the  ArcFace embeddings on the MOBIO database in terms of the  SAR. 
As this table shows, the irreversibility of MLP-Hash is comparable to that of the BioHash and IoM-GRP methods. However, IoM-URP protected templates are more difficult to invert using our adopted inversion technique.

\subsection{Recognition Accuracy Evaluation}\label{subsec:recognition_performance}
To evaluate the recognition accuracy of MLP-Hash, we considered two scenarios: the \textit{normal} scenario and the \textit{stolen token}  scenario. 
In the \textit{normal} scenario, which is the expected scenario for most cases, each user's key is assumed to be secret. 
However, in the \textit{stolen token} scenario (or briefly \textit{stolen} scenario), we assume that the impostor has access to the user's secret key and uses this key with the impostor's own unprotected template. 
To implement the \textit{stolen} scenario,  in the verification stage we 
used the same key as the genuine's key for other users in the database to generate their MLP-Hash templates.

Table \ref{tab:performance} compares the  MLP-Hash-protected, BioHash-protected, IoM-GRP-protected, IoM-URP-protected, and unprotected (baseline) templates of the SOTA face recognition models, in terms of True Match Rate (TMR) in the \textit{normal} and the \textit{stolen} scenarios on the MOBIO and LFW datasets. 
The threshold in each system is selected individually at an FMR of $10^{-3}$.  
As this table shows, in the normal scenario,  all the protection schemes achieve comparable performance on the MOBIO dataset. However, on the LFW dataset, IoM-URP clearly has the worst performance.  In the stolen scenario, IoM-GRP appears to perform the best across all three face recognition models and both evaluation datasets.

\begin{table}[tb]
	\centering
	\setlength{\tabcolsep}{3pt}
	\caption{Complexity comparison of template protection methods in terms of average  execution time (milliseconds). The results are reported as (mean$\pm$std) for 1000 different experimental trials.
	}\vspace{-1pt}
	\label{tab:complexity}
	\scalebox{.95}{
		\begin{tabular}[t]{@{} cccc @{}}
			\toprule[1pt]
			{\textbf{MLP-Hash}} & {\textbf{BioHash}}& {\textbf{IoM-GRP}} & {\textbf{IoM-URP}} \\
			\midrule[1pt]
			$61.9\pm0.5$ & $12.5\pm0.5$  & $77.6\pm0.2$ & $36.2\pm0.9$  \\ 
			\bottomrule[1pt]
		\end{tabular}
	}\vspace{-5pt}
\end{table}

\subsection{Discussion}\label{subsec:discussion}
Table \ref{tab:unlinkability}, Table \ref{tab:irreversibility}, and Table \ref{tab:performance} compare the unlinkability, irreversibility and recognition accuracy, respectively, of our proposed template protection method with the BioHash, IoM-GRP, and IoM-URP algorithms.
Table \ref{tab:complexity} also compares  the complexity of the aforementioned methods in generating protected templates from the ArcFace model in terms of average execution time (milliseconds) on a system equipped with an Intel(R) Core(TM) i7-7700K CPU @ 4.20GHz. 
Based on these results,
IoM-URP is the most irreversible algorithm, however it clearly has the worst performance in the normal scenario (which is the expected scenario in practice).
IoM-GRP has slightly better irreversibility than MLP-Hash, and its recognition accuracy is the best in most cases. However, it has the longest execution time amongst the studied protection methods.
BioHashing has comparable recognition accuracy  with MLP-Hash,  and has slightly worse irreversibility. However, BioHashing has the shortest execution time. All in all, our experiments show that  while all these template protection schemes have comparable unlinkability, there is a trade-off between irreversibility, recognition accuracy, and complexity.

\section{Conclusion}\label{sec:conclusion} 
In this paper, we proposed a new cancelable biometric template protection scheme, dubbed MLP-Hash, which uses a user-specific randomly-weighted multi-layer perceptron (MLP) with non-linear activation functions, followed by binarization of the output.
We evaluated the unlinkability, irreversibility and recognition accuracy of MLP-Hash as per the  ISO/IEC 30136 standard requirements, using SOTA face recognition models. 
Our protection method was found to satisfy these criteria to a high degree. 
In addition, we compared MLP-Hash with the BioHashing and IoM Hashing (IoM-GRP and IoM-URP) protection algorithms on the same SOTA face recognition systems, in terms of the recognition accuracy, unlinkability, and irreversibility criteria. 
Our experiments indicate that  while all these template protection schemes are almost unlinkable, there is a trade-off between irreversibility, recognition accuracy, and complexity.

\bibliographystyle{IEEEbib}
\bibliography{refs}
\end{document}